\documentclass[a4paper, 10pt, conference]{ieeeconf}      

\usepackage{FG2020}
\usepackage{times}
\usepackage{epsfig}
\usepackage{graphicx}
\usepackage{amsmath}
\usepackage{amssymb}
\usepackage{multirow}
\usepackage[outercaption]{sidecap}    

\IEEEoverridecommandlockouts                              
\usepackage[pagebackref=true,breaklinks=true,letterpaper=true,colorlinks,bookmarks=false]{hyperref}

\FGfinalcopy 

\def\Loss{\mathcal{L}}

\def\s{{\bf s}}
\def\Re{\mathbb{R}}

\def\eE{\mathbb{E}}

\title{\LARGE \bf A recurrent cycle consistency loss for progressive face-to-face synthesis}

\begin{document}
\ifFGfinal
\author{Enrique Sanchez$^{1,2,\dagger}$\thanks{$^\dagger$Work primarily done while at the University of Nottingham} and Michel Valstar$^2$ \\
    $^1$\,Samsung AI Center, Cambridge, UK \\
    $^2$\,Computer Vision Lab, University of Nottingham, UK \\
  }
  
\else
\author{Anonymous FG2020 submission\\ Paper ID \FGPaperID \\}
\pagestyle{plain}
\fi
\maketitle



\begin{abstract}
\noindent This paper addresses a major flaw of the cycle consistency loss when used to preserve the input appearance in the face-to-face synthesis domain. In particular, we show that the images generated by a network trained using this loss conceal a noise that hinders their use for further tasks. To overcome this limitation, we propose a ``recurrent cycle consistency loss" which for different sequences of target attributes minimises the distance between the output images, independent of any intermediate step. We empirically validate not only that our loss enables the re-use of generated images, but that it also improves their quality. In addition, we propose the very first network that covers the task of unconstrained landmark-guided face-to-face synthesis. Contrary to previous works, our proposed approach enables the transfer of a particular set of input features to a large span of poses and expressions, whereby the target landmarks become the ground-truth points. We then evaluate the consistency of our proposed approach to synthesise faces at the target landmarks. To the best of our knowledge, we are the first to propose a loss to overcome the limitation of the cycle consistency loss, and the first to propose an ``in-the-wild" landmark guided synthesis approach. Code and models for this paper can be found in \url{https://github.com/ESanchezLozano/GANnotation}. 
\end{abstract}

\section{Introduction}

\begin{figure}[t!]
\centering
  \includegraphics[width=0.93\columnwidth]{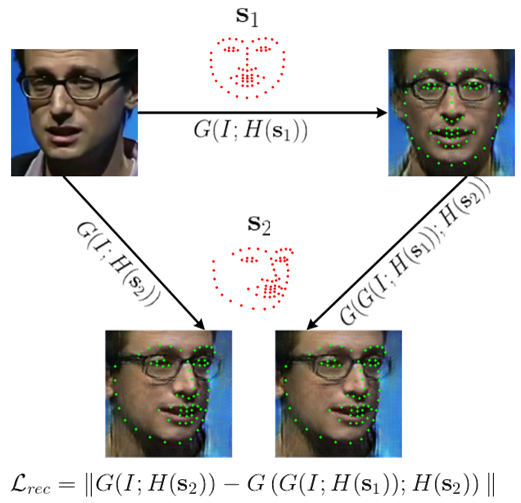}
  \caption{Applying a recurrent cycle consistency loss on generated images favours networks to produce similar results irrespective of their path to the target; either directly to $\s_2$ or via $\s_1$. $G$ is the generator function, $I$ is the input image, and $H(\s_t)$ are the heatmaps defined by the target shape $\s_t$. All images but the one without the landmarks are synthetic.}
  \label{fig:triplelossabstract}
\end{figure}

\begin{figure*}[t!]
    \centering\includegraphics[width=1\linewidth]{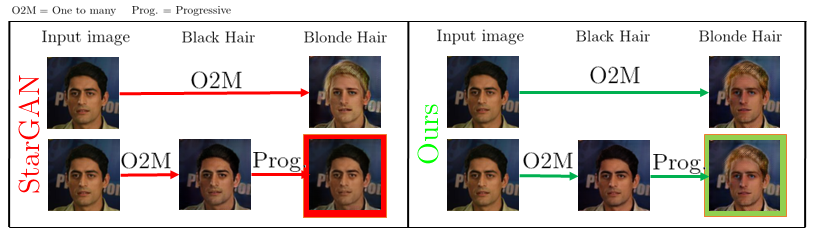}\par
  
  \caption{This paper illustrates the drawback of the cycle consistency loss to preserve the input features. The left figure represents the use of StarGAN~\cite{Choi2018}. O2M stands for one-to-many, whereby attributes are placed directly onto the input images. Progr. stands for progressive, whereby the output of the network is used to generate the next attribute. In the top row the ``Blonde Hair" attribute is directly, and correctly, placed on top of the input image. In the bottom row, the input image is first passed through the network to generate the ``Black Hair" attribute, and then passed through the network to generate the ``Blonde Hair" attribute. StarGAN fails to generate the new attribute. In the right side, we can see two advantages of our proposed loss: 1) the attribute ``Blonde Hair" is correctly placed after a first pass on the network, and 2) the quality of the generated images in the O2M approach improves that of StarGAN. We validate this qualitatively and quantitatively in Sec.~\ref{sec:loss}. }
  \label{problem}
\end{figure*}

\noindent Recent advances in Generative Adversarial Networks (GANs~\cite{Goodfellow2014}) have found a broad range of applications in the domain of face synthesis or face-to-face translation~\cite{Huang2017,Choi2018,Di2018,Kossaifi2018,Pumarola2018}, where the goal is to translate, or place, a set of attributes onto an input face. Herein, we refer to ``translating an attribute" as modifying the hair colour~\cite{Choi2018}, the pose~\cite{Zhao2017}, the expression~\cite{Pumarola2018, Wang2018}, or even the landmarks, as proposed in this paper~(see Fig.~\ref{fig:triplelossabstract}). An additional goal of face-to-face translation methods is to preserve all features in a given face other than those set as the target\footnote{This concept is often referred to as ``identity preserving". However, it depends on the target task whether preserving identity is even possible, as e.g. changing shape or appearance is a task that naturally destroys identity.}. 
The majority of recent approaches rely on the use of a cycle consistency loss~\cite{Zhu2017} (also known as the reconstruction loss) to help a network preserve relevant input features. To compute the cycle consistency loss, an image is first generated using a specific target attribute. Then, the output of the network and the ``reverted" attribute target are sent to the network. The reverted attribute refers to the ground-truth attribute of the original input image. The cycle consistency loss measures the difference between this reconstruction and the input image. It is a desired property for the generated images to be reusable, i.e. to follow a similar probability distribution as that of the input images. However, while the generated images can be said to be photo-realistic, the distributions generated by the current state of the art differ in an important way from the corresponding input domain. Upon closer inspection, we can reveal an interesting phenomenon: when the generated images are re-introduced to the network with a new set of target attributes, the network yields poor results, and occasionally even fails to produce photo-realistic images. In particular, we observe that when using the state of the art StarGAN network for facial attribute setting~\cite{Choi2018}, if the first output of the network is re-used to generate a second attribute, the input image is recovered, no matter what the second attribute is. This effect is illustrated in Fig.~\ref{problem}. In other words, the network leaves a footprint in the generated image, not perceptible to the human eye, but that is evident when the generated image is re-introduced to the network. That is, the generated images cannot be reused for further tasks. Consider the following example goal: \textit{Can we use a network to convert the hair of a given person in an image from blonde to brown, and then use another network to modify their corresponding facial expression?} Without addressing the flaws of the methods based on the cycle consistency loss, the answer is no. 

\indent This paper presents an approach to tackle this problem by introducing a new consistency loss, coined \textit{recurrent cycle consistency loss} (Fig.~\ref{fig:triplelossabstract}). This loss aims to bridge the gap between the input and target distributions by imposing that any generated image has to be the same no matter if it is targeted by the network directly in one step or through a sequence of two or more steps. To validate the generalisation of this loss, we first introduce it into the training of StarGAN, and show its effectiveness. After retraining the StarGAN network with our loss, we observe a systematic improvement in the quality of the generated images, both in the one-to-many and the progressive approaches. (Fig.~\ref{problem}, right). \\
\indent Next, we present our novel approach to unconstrained landmark guided face-to-face synthesis, which we name \textit{GANnotation}, and prove again the effectiveness of our introduced loss. GANnotation translates a given face to a set of target landmarks, given to the network in the form of heatmaps. We show that the target landmarks become the ground-truth points at the generated images. To the best of our knowledge, our GANnotation is the first network that allows synthesising faces in a wide range of poses and expressions. An example is depicted in Fig.~\ref{fig:triplelossabstract}, where the input image $I$ is translated into the target point configurations $\s_1$ (and $\s_2$). In short, the contributions of this paper can be summarised as follows:
\begin{itemize}
\item We are the first to propose a \textbf{recurrent cycle consistency loss} to bridge the gap between the distributions of the input and generated images. This enables the training of networks that not only reproduce photo-realistic images, but are also suitable for its use in combination with other networks. 
\item We propose \textbf{GANnotation}, the first network that applies a face-to-face synthesis with simultaneous change in pose and expression, whereby the given landmarks correspond to the ground-truth points in the generated images. 
\end{itemize}

\section{Related Work}
\label{sec:related}
\noindent Generative Adversarial Networks (GANs)~\cite{Goodfellow2014} are a powerful tool in many Computer Vision disciplines, such as image generation~\cite{Radford2015}, style transfer~\cite{Johnson2016}, or super-resolution~\cite{Ledig2017}.
In the context of image to image translation~\cite{Isola2017,Zhu2017}, GANs are composed of a generator that aims to reproduce the target domain, and a discriminator that tells whether the output of the generator is close to the target distribution or not. Both are learnt simultaneously using the minimax strategy. Since the introduction of GANs, many improvements on  adversarial learning have been proposed, including the Least-Squares GAN~\cite{Mao2017}, the Wasserstein GAN~\cite{Gulrajani2017,Arjovsky2017}, the Geometric GAN~\cite{Lim2017}, or  Spectral Normalisation~\cite{Wang2018non,Zhang2018,Miyato2018}, however there is no consensus as to which exhibits a systematic improvement~\cite{Arjovsky2017b,Lucic2018}. \\
\indent Reports on works suggesting improvements to the state of the art in GANs are often applied to the face domain. We review those that we consider the closest to our proposed approach, which aims to generate faces conditioned to attributes, landmarks, or expressions. There are works that proposed to do face frontalisation (\textbf{TP-GAN}~\cite{Huang2017}, \textbf{FF-GAN}~\cite{Yin2017}), profile face synthesis (\textbf{DA-GAN}~\cite{Zhao2017}), or multi-view image generation (\textbf{CR-GAN}~\cite{Tian2018}). However, these methods do not allow the synthesis of customised expressions or poses, so while CR-GAN can generate $9$ different views, these can not include synthesised expressions in addition. We will show how our GANnotation \emph{can} perform both tasks with a landmark-guided synthesis. In principle other tasks based on face shape could be added, such as changing a thin face to a round face or thin lips to full lips.\\ 
\indent Some other works use geometric information to help synthesise expressions or pose. Our GANnotation is the first network that allows the synthesis of both. For instance, \textbf{CMN-Net}~\cite{Wang2018} is a landmark guided smile generator that generates frontal images under different expressions, supported by a recurrent neural network to preserve spatial consistency in the landmark generation. However, this method is limited to frontal faces, and the input image is expected to be neutral. Also, the \textbf{GC-GAN}~\cite{Qiao2018} is a geometry-aware method that is used to synthesise images from landmarks, where these are meant to display expressions. However, this method does not account for changes in pose. In contrast, the \textbf{CAPG-GAN}~\cite{Hu2018} applies pose-specific face rotation, where the input and target pose are encoded in sets of five heatmaps each, so that the network can perform attention. The five points are meant to capture head pose, which as an unwanted side-effect removes the network's ability to perform expression synthesis. The problem of expression synthesis was also approached by \textbf{GANimation}~\cite{Pumarola2018}, where the generated images undergo a translation in the displayed expression. Rather than using geometric information, GANimation relies on an auxiliary expression loss. The use of a conditional loss is also used in \textbf{StarGAN}~\cite{Choi2018}. StarGAN is a multi-domain face-to-face synthesis network that allows the synthesis of different facial attributes, such as ``Blonde Hair", as well as expressions. Like the majority of networks mentioned above, StarGAN is limited to frontal poses only.   \\
\indent The majority of the methods mentioned above rely on the use of a cycle consistency loss to preserve the input features. In particular, the cycle loss is key to the success of \textbf{StarGAN}~\cite{Choi2018}. As shown in Fig.~\ref{problem}, this loss limits existing methods to  one-to-one mappings, and renders images that are unable to be used as the basis for generating further images. These methods might leave a neutral face to a given expression~\cite{Pumarola2018}, or a non-frontal face to a frontal one~\cite{Huang2017}. In either case, the network is not required to perform more than one forward pass from a given image. Thus, the cycle consistency loss is applied to preserve the input appearance. While this yields impressive results, it causes a mismatch between the input and target distributions, when a desired property would be to actually make them match. \\
\indent Finally, it is worth mentioning that there exist other works that have proposed landmark-guided synthesis from a random seed, but without the aim to preserve the input features. \textbf{GP-GAN}~\cite{Di2018} and \textbf{GAGAN}~\cite{Kossaifi2018}, are two good examples. Both the GP-GAN and GAGAN generate random faces, liying out of the domain of face-to-face synthesis. \\
\indent As we shall see, our proposed network is the first to fill the gap of ``in-the-wild" landmark-guided face to face synthesis, where the generated images can be reused for subsequent image generation tasks.

\section{Recurrent Cycle Consistency Loss}
\noindent In this Section, we illustrate a general framework of face-to-face synthesis conditioned to a target attribute $\s_t$\footnote{$\s_t$ is chosen as a convenient reference to a shape, which is the input attribute in our GANnotation described in Sec.~\ref{sec:ganno}} where a cycle consistency loss is used to help the network preserve the input features. The goal is to learn a network $G$ that, given an image $I$ and target attribute $\s_t$, is capable of generating an image that possesses that attribute, and that preserves the input appearance in all aspects other than those required by the target attribute. 

\label{sec:loss}
\subsection{Notation}
\label{sec:notation}
\noindent Let $I \in \mathcal{I}$ be a $w \times h$ pixels face image. $\mathcal{I}$ represents the space of images of size  ${w \times h}$. Let $\mathcal{H}$ represent the space of encoded attributes. The \textit{generator} is a function $G: \mathcal{I} \times \mathcal{H} \rightarrow \mathcal{I}$, that receives as input an image $I$ and an attribute representation $H(\s_t) \in \mathcal{H}$ encoding the target attribute $\s_t$, and outputs the generated image. Without loss of generality, we define the estimated image $\hat{I}$ as:
\begin{equation}
\hat{I} = G\left( I ; H(\s_t) \right)
\end{equation}
where $;$ indicates that $I$ and $H$ are concatenated. In the StarGAN setup, $H$ is a one-hot vector representing the target attribute $\s_t$, reshaped to replicate the input spatial resolution. In our proposed approach, $H(\s)$ is a heatmap representation of a target shape $\s_t$. In both cases,  $H(\s) \in \Re^{n \times w \times h}$, with $n$ the number of attributes in StarGAN, or points in our GANnotation (see Sec~\ref{sec:ganno}). In the adversarial learning framework, a \textit{discriminator} is defined as a function $D$ that receives as input an estimated image $\hat{I}$, or a real image $I$, and aims to label them as real or fake. The training requires a set of $N$ images and labelled attributes $\{I, \s\}_{i=1:N}$. The learning is accomplished using a minimax strategy, where the generator and discriminator are updated in an iterative way. For the sake of clarity, we assume that the total cost to be optimised is composed of an adversarial loss, and a set of auxiliary losses, which are omitted here. Further details on the training of GANnotation can be found in Sec.~\ref{sec:ganno}, whereas the details of StarGAN training can be found in~\cite{Choi2018}. 
\subsection{Cycle Consistency Loss} 
\label{ssec:cycleloss}
\noindent In the context of face to face synthesis, the cycle consistency loss is used to preserve the input features in the generated images. In~\cite{Pumarola2018,Choi2018} this loss is also referred to as identity or reconstruction loss. In practice, the input image is paired with an attribute $\s_i$. The cycle consistency loss applies the original attribute $\s_i$ to the generated image for the target attribute $\s_t$, and measures the distance w.r.t. the input $I$: 
\begin{equation}
\Loss_{cyc} = \| G\left( G( I ; H(\s_t) ) ; H( \s_i ) \right) - I \|^2.
\label{selfloss}
\end{equation}
In the StarGAN, $\s_i$ refers to the actual label of the input image. In our GANnotation shown below, $\s_i$ refers to the ground-truth points of the input image. \\ 
\indent As shown above, the drawback of this loss is that it encourages the network to imprint a footprint on the images that impedes its further use. We validated this issue empirically. To do so, we re-use the StarGAN~\cite{Choi2018} implementation provided with the author's trained model. Recall that one of the key aspects behind the success of StarGAN relies on the use of the cycle consistency loss shown in Eqn.~\ref{selfloss} to preserve the content of the input images. The original StarGAN model was trained on the Celeb-A dataset~\cite{Liu2015}, and it applies to a given face a set of attributes, namely ``Black Hair'', ``Blonde Hair'', ``Brown Hair'', ``Gender'', and ``Age''. The attributes of ``Gender'' and ``Age'' have to be understood as generating the opposite attribute to the one given in the input image. 
\indent Using the pre-trained model, we generated for each image all the attributes in a \textit{progressive} way, i.e. taking as input the output of the network after generating the previous attribute. The ordering of attributes is the same as shown above. In other words, for each image, the model is used to first generate the ``Black Hair" attribute. Then, the output of the network is used to generate the ``Blonde Hair" attribute, etc. In addition to the results shown in Fig.~\ref{problem}, a set of qualitative results is shown in the left side of Fig.~\ref{fig:stargan}, where the results for the one to many approach (i.e.  always using the input image as the source to generate the target attributes) are compared with those yielded by the network in the progressive approach. The figures clearly show that after the second pass, when the ``Blonde Hair" attribute is set as target, the output of the network recovers the input image, i.e. it fails to produce the target attribute. More test images are added as Supplementary Material. We also observed this effect in our GANnotation without our recurrent cycle consistency loss (Sec.~\ref{gannot}, Fig.~\ref{fig:onto}).   
\subsection{Recurrent Cycle Consistency Loss}
\label{ssec:reccycleloss}
\begin{figure*}[t!]
\begin{tabular}{|p{0.1cm}|*{6}{c}|*{6}{c}|} \hline 
   & \multicolumn{6}{c|}{StarGAN without recurrent cycle consistency loss}  & \multicolumn{6}{c|}{StarGAN with recurrent cycle consistency loss} \\ \hline
   & \hspace{7pt} \scriptsize{Input} & \scriptsize{Black Hair} & \scriptsize{Blonde Hair} & \scriptsize{Brown Hair} & \scriptsize{Gender} & \scriptsize{Age} & \hspace{7pt} \scriptsize{Input} & \scriptsize{Black Hair} & \scriptsize{Blonde Hair} & \scriptsize{Brown Hair} & \scriptsize{Gender} & \scriptsize{Age} \\
  \rotatebox{90}{\hspace{2pt} \scriptsize{Progressive | O2M}} & \multicolumn{6}{c|}{\includegraphics[width=0.95\columnwidth]{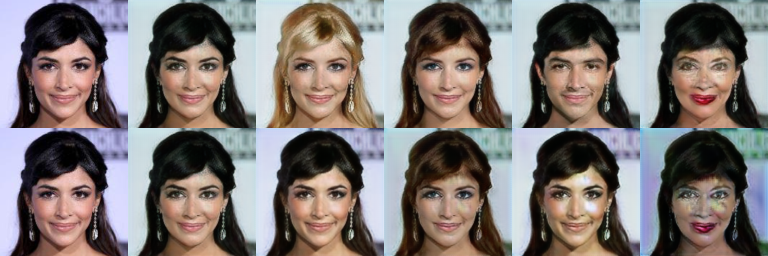}} &  \multicolumn{6}{c|}{\includegraphics[width=0.95\columnwidth]{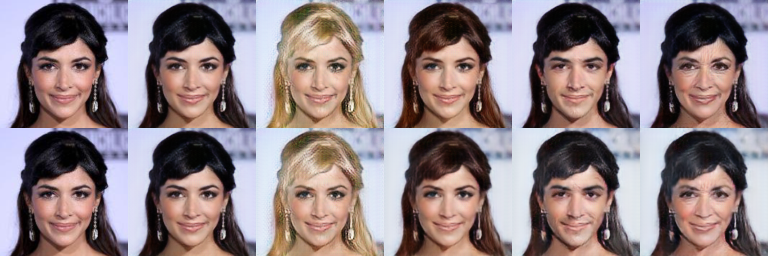}} \\ 
 \rotatebox{90}{\hspace{2pt} \scriptsize{Progressive | O2M}}  & \multicolumn{6}{c|}{\includegraphics[width=0.95\columnwidth]{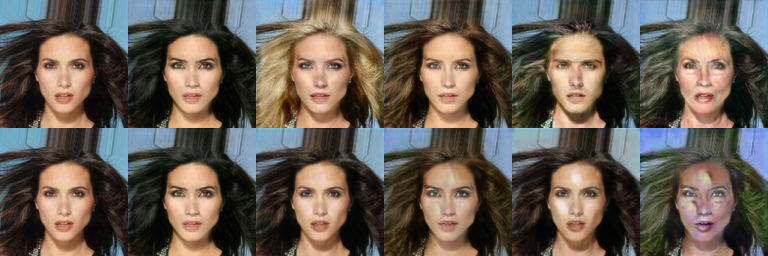}} & \multicolumn{6}{c|}{\includegraphics[width=0.95\columnwidth]{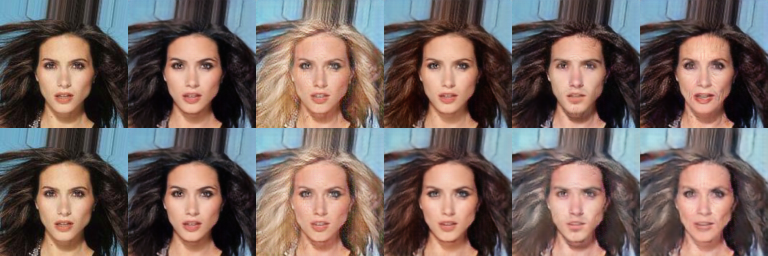}} \\ 
  \rotatebox{90}{\hspace{2pt} \scriptsize{Progressive | O2M}}  &
  \multicolumn{6}{c|}{\includegraphics[width=0.95\columnwidth]{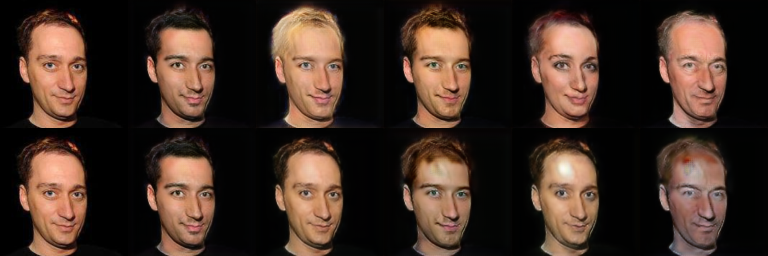}} & \multicolumn{6}{c|}{\includegraphics[width=0.95\columnwidth]{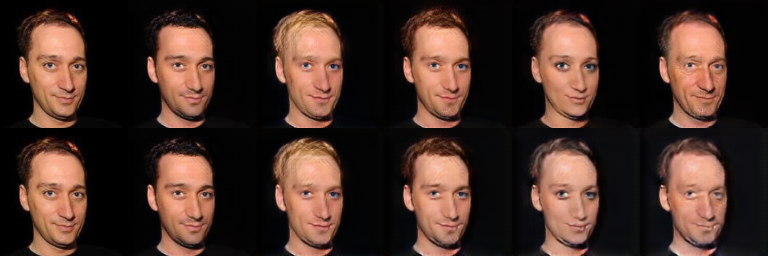}} \\ \hline
   \end{tabular}
 
  \caption{Comparison between the original StarGAN~\cite{Choi2018} (left) w.r.t. a StarGAN trained with the recurrent cycle consistency loss introduced in this paper (right). In each example, the first row corresponds to the one-to-many approach, where the input image is always used to generate the target attribute. The second row shows the results after using the output of the network as input for the next target attribute. More examples can be found in the Supp Material.}
  \label{fig:stargan}
\end{figure*}
\noindent We have validated that, when using the Cycle Consistency Loss, the network recovers the input image \textit{no matter what further target is considered}, when we expect this to only happen with the further target set as the inverse of the original target. That is to say, the network translates images into a domain that encodes the input image along with the output. We conjecture that this problem has so far remained undiscovered due to the fact that existing works set a neutral-to-expression synthesis goal rather than expression-to-expression, which means that the input and output spaces do not need to overlap. However, we want the network to produce photo-realistic images that are also reusable, and therefore the input and output domains need to be similar. \\
\indent In order to solve this problem, we propose a recurrent cycle consistency loss, which aims to enforce the network to ``pull-out" the encoded information so that it remains invisible after a second pass. In particular, if an image is first translated into an attribute $\s_t$, and then further translated into an attribute $\s_n$, the output should be similar to that given by the network if the original image was directly translated into the attribute $\s_n$. Given the input image $I$, and target attribute $\s_t$, the output of the generator is $\hat{I} = G( I ; H(\s_t) )$. Now, we observe that translating $I$ and $\hat{I}$ to another target attribute $\s_n$ should result in similar outputs. That is to say, we want $G( \hat{I} ; H(\s_n) )$ to be similar to $G(I ; H(\s_n))$. The recurrent cycle consistency loss is thus defined as:
\begin{equation}
\Loss_{rec} = \| G( \hat{I} ; H(\s_n) ) - G( I ; H(\s_n) ) \|^2
\end{equation}
The overall idea of the recurrent cycle consistency loss is depicted in Fig.~\ref{fig:triplelossabstract}. \\
\indent In order to validate the contribution of our proposed loss, we re-trained the original StarGAN network with our recurrent cycle consistency loss, exactly under the same conditions as those of the original network. The contribution of the recurrent cycle consistency loss was balanced with that of the cycle consistency loss. Then, we tested the trained model repeating the process described above. The qualitative results are shown in the right half of Fig.~\ref{fig:stargan}.

\begin{table}[t!]
\centering
\caption{FID scores for the CelebA dataset. O2M stands for the one-to-many approach. }
\label{table:fid_stargan}

\begin{tabular}{l|l|l|l|l|l|}
 \hline
\multicolumn{1}{|l|}{}                     & Black & Blon. & Brown & Gen. & Age   \\ \hline
\multicolumn{1}{|l|}{StarGAN O2M}    & 21.4 & 28.6 & 21.6 & 22.0 & 22.8  \\ \hline
\multicolumn{1}{|l|}{Ours O2M}      & 20.7 & 25.6 & 17.8 & 18.0 & 16.2       \\ \hline
\multicolumn{1}{|l|}{StarGAN Progr.} & 21.3 & 21.0  & 28.2 & 32.3 & 39.2  \\ \hline
\multicolumn{1}{|l|}{Ours Progr.}   & 20.7 & 27.3 & 27.1 & 33.2 & 35.6     \\ \hline
\end{tabular}
\end{table}
\subsection{Quantitative evaluation}
\noindent We observe that adding our loss to the StarGAN training not only qualifies the network to produce results in a progressive way, but also improves the quality of the generated images in the one-to-many approach. This is mainly due to the fact that our loss is cleaning the noise introduced by the cycle consistency loss. To show this, in addition to the results shown in Fig.~\ref{fig:stargan} and the Supplementary Material, we measured the FID (Fr\'{e}chet Inception Distance,~\cite{Heusel2017}) between the input images of the test partition of CelebA ($2000$ images), and the generated ones. The FID uses the second order statistics measured from the last layer of the Inception Network of~\cite{Szegedy2015}, and serves as a similarity measure. The results are shown in Table~\ref{table:fid_stargan}. It can be seen that in the one-to-many approach, the FID scores are slightly different for each of the attributes, for all the methods. This difference comes from the different statistics that appear in the CelebA dataset (e.g. ``Black Hair'' and ``Blonde Hair'' correspond to $25\%$ and $14\%$ of the images, respectively). When correctly placing the attributes Black and Blonde Hair, one expects the generated images of the former to be statistically closer to the original images than those of the latter, which explains the lower FID score for the ``Black Hair" attribute w.r.t. that of the ``Blonde Hair". This difference can be observed for all the rows shown in Table~\ref{table:fid_stargan} but the one corresponding to StarGAN in the progressive approach. Provided that StarGAN fails to correctly place the ``Blonde Hair'' attribute in the progressive approach, and considering that the original images are recovered due to the consistency constraint, the generated images are statistically close to the original ones, and hence the FID for the ``Blonde Hair'' attribute ($21.0$) is even be lower than the FID for the ``Black Hair'' attribute ($21.3$). This lower FID needs to be analysed alongside the visual evidence provided in the Supplementary Material, to validate that it corresponds to a failure in using the original StarGAN in a progressive way.

\section{GANnotation}
\label{sec:ganno}
\begin{figure}[h!]
\centering
  \includegraphics[width=0.97\columnwidth]{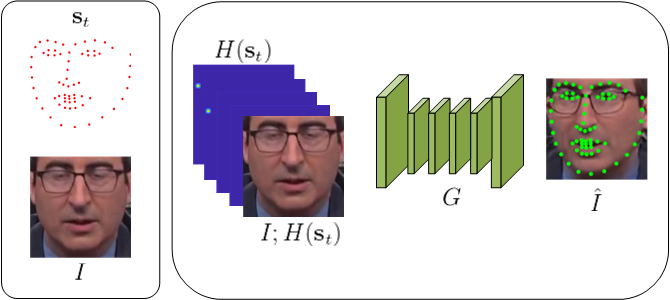}
  \caption{Proposed approach: We are given an input image $I$, and a set of target points $\s_t$. The points are encoded as a set of heatmaps $H(\s_t)$ and concatenated with the input image $I;H(\s_t)$. The concatenated volume is sent to the generator $G$ to produce the image $\hat{I}$, we overlay the target points on the generated image to illustrate the main task of the network.}
  \label{fig:network}
\end{figure}

\noindent We now introduce GANnotation, our framework to generate (synthesise) a set of person-specific images driven by a set of landmarks, so that these become the ground-truth landmarks in the generated image. Contrary to previous works, we want our network to allow for simultaneous changes in both pose and expression. To the best of our knowledge, this is the first work that directly permits changes in pose and expression simultaneously. An overall description of our proposed approach is depicted in Fig.~\ref{fig:network}.

\subsection{Architecture}
\label{ssec:architecture}
\noindent The generator is adapted from the architecture successfully proposed for the task of neural transfer~\cite{Johnson2016}, and later adapted to the image-to-image translation task~\cite{Isola2017,Zhu2017, Sanchez19}. This architecture has also proven successful for the task of face synthesis~\cite{Pumarola2018,Ma2017,Hu2018}, and basically consists of two spatial downsampling convolutions, followed by a set of residual blocks~\cite{He2016}, and two spatial upsampling blocks with $1/2$ strided convolutions. We encode the target attribute (the target locations), using heatmaps, one per point, each being a unit Gaussian centred at the corresponding landmark. The generator is then modified to account for the $ 3 + n $ input channels, defined by the RGB input image and the heatmaps corresponding to the target landmark locations. As in~\cite{Pumarola2018}, we adopt a mask-based approach, by splitting the last layer of the generator into a colour image $C$ and a mask $M$. The output of the generator is thus defined as:
\begin{equation}
\hat{I} = (1 - M) \circ C + M \circ I,
\end{equation}
where $\circ$ represents an element-wise product. Without loss of generality, we will refer to $\hat{I}$ as the output of the generator. Further details of the network can be found in the Supplementary Material. 

The discriminator is adopted from the PatchGAN~\cite{Isola2017,Zhu2017} network, with several convolution-based downsampling blocks, each increasing the number of channels to $512$, and followed by a LeakyReLU~\cite{Maas2013}. For an input resolution of $128\times 128$ this network yields an output volume of $4 \times 4 \times 512$, which is forwarded to a FCN to give a final score.

\subsection{Training}
\label{ssec:training}
\noindent The loss function consists of five terms: an adversarial loss, a pixel loss, the cycle and recurrent cycle consistency losses, a perceptual loss, and a total variation regularisation. 

\subsubsection{Adversarial loss} 
\label{ssec:adversarial}
\noindent We adopt the hinge adversarial loss proposed in~\cite{Lim2017}, which is shown to require fewer updates in the discriminator per update in the generator, thus enabling faster learning~\cite{Zhang2018,Miyato2018}. The loss for the discriminator is defined as:
\begin{equation}
    \mathcal{L}_{adv} = -\eE_{\hat{I}} [ \min(0,-1+D(\hat{I}))] -  \eE_{I} [\min(0,-D(I)-1) ]
\end{equation}
whereas the loss for the generator is defined as:
\begin{equation}
\mathcal{L}_{adv} = -\eE_{I} [  D(\hat{I}) ].
\end{equation}

\subsubsection{Pixel Loss} 
\label{ssec:pixeloss}
\noindent In order to make the network learn the target representation, we use a pixel reconstruction loss. In particular, we assume that, for a given input image $I$, and target points $\s_t$, there is an available ground-truth image $I_t$, that can be used to compute a pixel reconstruction loss:
\begin{equation}
\Loss_{pix} = \| G(I ; H(\s_t)) - I_t \|_2^2.
\end{equation}
In order to have access to paired data, we construct our training set from video sequences, from which the annotations are available for each frame. This gives a fairly large amount of pairwise combinations. In addition, we can augment the training set with still images. We can see that each image can be paired with a rigid transformation of itself. If we are given an image $I$ with ground-truth points $\s_i$, we can generate a target shape $\s_t$, and its corresponding ground-truth image $I_t$, by applying a rigid transformation (rotation, scale, and translation) to both $I$ and $\s_i$.

\subsubsection{Auxiliary Losses}
\label{ssec:auxiliary}
\noindent In order to enforce the network to preserve the input features, we use the cycle consistency loss presented in Sec.~\ref{ssec:cycleloss}, and the recurrent cycle consistency loss shown in Sec.~\ref{ssec:reccycleloss}. In addition, and in order to provide the network with the ability to generate subtle details, we follow the line of recent approaches in super resolution and style transfer~\cite{Ledig2017,Bulat2018}, and use the perceptual loss defined by~\cite{Johnson2016}. The perceptual loss enforces the features at the generated images to be similar to those of the real images when forwarded through a VGG-19~\cite{Simonyan15} network. The perceptual loss is split between the feature reconstruction loss and the style reconstruction loss. The feature reconstruction loss is computed as the $l_1$-norm of the difference between the features $\Phi^{l}_{VGG}$ computed at the layers $l = \{\mathtt{relu1\_2}, \mathtt{relu2\_2}, \mathtt{relu3\_3},\mathtt{relu4\_3}\}$ of the input and generated images. The style reconstruction loss is computed as the Frobenius norm of the difference between the Gram matrices, $\Gamma$, of the output and target images, computed from the features extracted at the $\mathtt{relu3\_3}$ layer. The perceptual loss is referred to as $\Loss_{pp}$. Finally, we also apply a \textit{total variation regularisation}, $\Loss_{tv}$~\cite{Aly2005,Johnson2016}, which encourages smoothness in the generated images.

\subsubsection{Full loss}
The full loss for the generator is defined as:
\begin{equation}
    \Loss = \lambda_{adv} \Loss_{adv} + \lambda_{pix} \Loss_{pix} + \lambda_{cyc} \Loss_{cyc}  
    + \lambda_{rec} \Loss_{rec} + \lambda_{tv} \Loss_{tv},
\end{equation}
where, in our set-up, $\lambda_{adv} = 1$, $\lambda_{pix} = 10$, $\lambda_{cyc} = 100$, $\lambda_{rec} = 100$, and $\lambda_{tv} = 10^{-4}$. The hyper-parameters were found by grid-search, and can vary according to the data.

\section{Experiments}
\noindent All the experiments are implemented in PyTorch~\cite{Paszke2017}, using the Adam optimiser~\cite{Kingma2015}, with $\beta_1 = 0.5$ and $\beta_2 = 0.9$. The input images are cropped according to a bounding box defined by the ground-truth landmarks with an added margin of 10 pixels each side, and then re-scaled to be 128x128. The model is trained for 30 epochs, each consisting of $10,000$ iterations, which takes approximately $24$ hours to be completed with two NVIDIA Titan X GPU cards. The batch size is $16$, and the learning rate is $10^{-4}$.

\subsection{Training Datasets}
\label{data}
\noindent To train our GANnotation, we use a set of videos from the training partition of the 300VW~\cite{Shen15}, which is composed of annotated videos of $50$ people. For each video, we randomly chose a set of $3000$ pairs of images. In addition, we use the public partition of the BP4D dataset~\cite{Zhang2014bp4d,Valstar2015}, which is composed of videos of $40$ subjects performing $8$ different tasks. For each of the BP4D videos, we select $500$ pairs of images. As mentioned in Sec.~\ref{ssec:pixeloss}, we augment our training set with still images. In particular, we use a subset of $\sim$8000 images collected from datasets that are annotated in a similar fashion to that of the 300VW. We use Helen~\cite{Le12}, LFPW~\cite{Belhumeur13}, AFW~\cite{Zhu12}, IBUG~\cite{Sagonas13}, and a subset of MultiPIE~\cite{Gross10}. To ensure label consistency across datasets we used the facial landmark annotations provided by the 300W challenge~\cite{Sagonas13}. We apply data augmentation during training, by randomly rotating and scaling both the images and the corresponding ground-truth points. 

\subsection{Impact of recurrent cycle consistency loss}
\noindent We first validate the contribution of our new loss in the context of GANnotation. To do so, we train two models under the same conditions, with, and without the proposed loss. At test time, we use a set of images from the test partition of 300VW~\cite{Shen15} for which there are available points. Each image is first frontalised using the given landmarks (see Section~\ref{gannot} for further details), and then sent to a pose-specific angle. The results are shown in Fig.~\ref{fig:onto}, where the odd rows correspond to the images generated by a model trained with the proposed loss and the even rows represent the images generated by a model trained without it. We show how after the first map, both images look alike, being similar to the input image. However, after the second pass, the generator trained without the recurrent cycle consistency loss recovers the input images, with subtle changes in the contrast. This effect is not occurring with the images generated by the network trained with the proposed loss, where the images are correctly mapped. We also show how the network produces similar results after the first pass.
\subsection{Qualitative evaluation}
\label{gannot}
\begin{figure}[h!]
\centering
  \includegraphics[width=0.45\columnwidth]{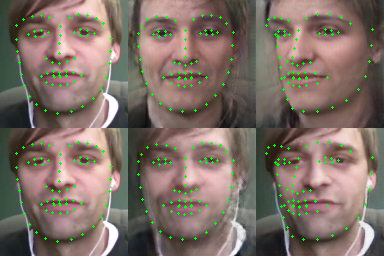} 
  \includegraphics[width=0.45\columnwidth]{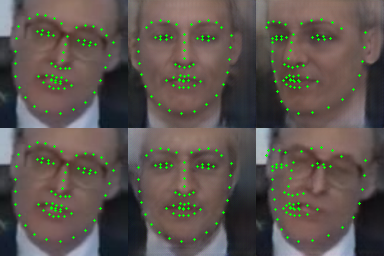}\\
  \includegraphics[width=0.45\columnwidth]{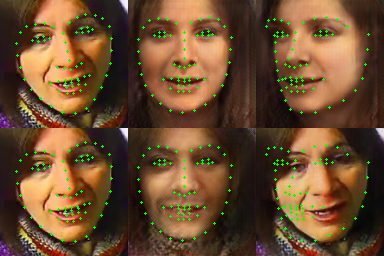}
  \includegraphics[width=0.45\columnwidth]{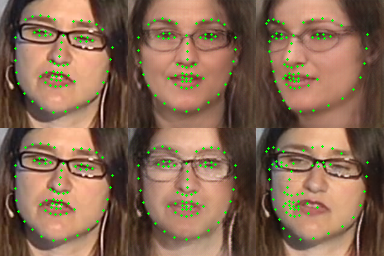} \\
  \includegraphics[width=0.45\columnwidth]{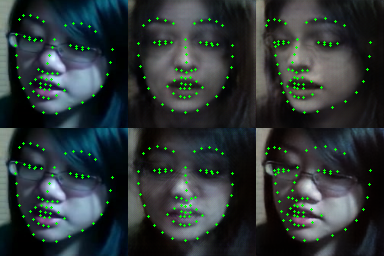}
  \includegraphics[width=0.45\columnwidth]{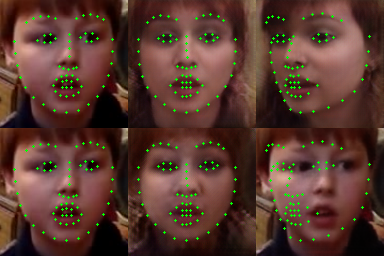} 
  \caption{Results of a model trained with our proposed loss (top rows) and a model trained without it (bottom rows). The left, image corresponds to the input image, the middle image corresponds to the output of the network after the first pass, and the right image corresponds to the output of the network after a second pass. The points represent the target location. \vspace{-15pt}}
  \label{fig:onto}
\end{figure}

\begin{figure*}[t!]
\centering
  \includegraphics[width=2\columnwidth]{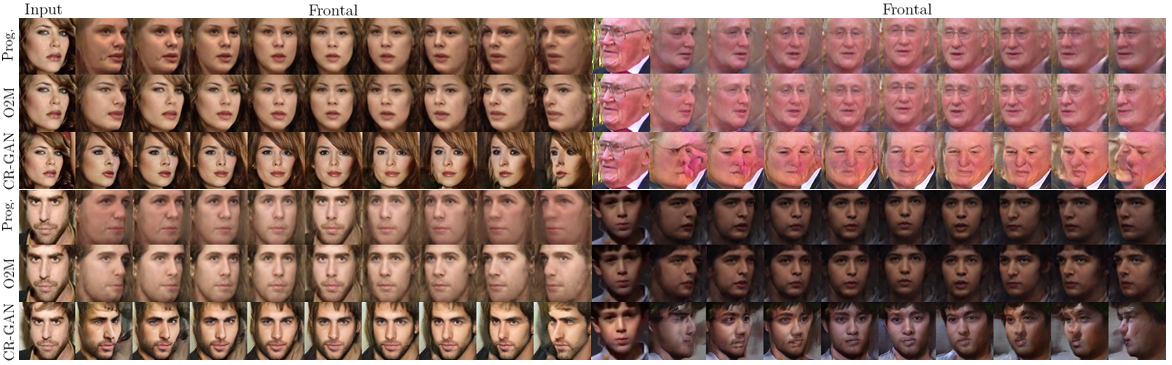}
  \caption{Landmark-guided multi-view synthesis and comparison with CR-GAN~\cite{Tian2018}. The first row corresponds to a progressive image generation, whereby the input image (leftmost) is first frontalised (``Frontal"), and then sent progressively to the corresponding views (i.e. left and right). The second row corresponds to a one-to-many mapping. The third row corresponds to the CR-GAN results. The two examples on the left images correspond to the images accompanying the code of CR-GAN. Our method yields realistic results despite the tight cropping, different to that used to train our GANnotation. On the right, two examples extracted from the 300VW test partition, cropped according to the landmarks, where CR-GAN fails to produce photo-realistic results.} 
  \label{fig:comparison_1}
\end{figure*}

\noindent We now evaluate the consistency of our GANnotation for the task of landmark-guided face synthesis. In order to compare our GANnotation w.r.t. the most recent works, we apply a landmark-guided multi-view synthesis, and compare our results against the publicly available code of \textbf{CR-GAN}~\cite{Tian2018}. We compare our method in the test partition of the 300VW~\cite{Shen15}. To generate pose-specific landmarks, we use a shape model trained on the datasets described in Section~\ref{data}. The shape model includes a set of specific parameters that allow manipulating the in-plane rotation, as well as the view angle (pose). Using the shape model, we first remove both the in-plane rotation and the pose, resulting in the frontalised image given in the middle column of each block of faces in Fig \ref{fig:onto}. Then, the pose specific parameter is manipulated to generate the synthetic poses shown in the left and right columns w.r.t. the frontalised face. In addition, when generating the pose-specific landmarks, we randomly perturb the expression related parameters, so as to generate different faces. The results are shown in Fig.~\ref{fig:comparison_1} for both the \textit{progressive} image generation (first row of each block), and the one-to-one mapping (second row). We compare the results w.r.t. those given by the CR-GAN model (third row). 
\begin{table}[h!]
\centering
\caption{Overall performance on the original and generated videos.}
\label{fig:testA}
\begin{tabular}{c|c|c|}
\cline{2-3}
                          & Generated & Original \\ \hline
\multicolumn{1}{|c|}{Area Under the Curve (AUC)} & 0.684     & 0.475    \\ \hline
\end{tabular}
\end{table}
\vspace{-0.5cm}
\subsection{Quantitative evaluation}
\label{ssec:qual_ganno}
\noindent We evaluated the correctness of the placed landmarks in a set of generated videos. To generate the videos, we used as target the landmarks corresponding to the videos on the most challenging category of the 300VW test partition. We use a random image as input to each video. Then, we used a state of the art face tracker~\cite{Sanchez16, Sanchez2018} to detect the landmarks both in the original and in the generated images. The Area Under the Curve is shown in Table~\ref{fig:testA}. Several of the original and generated videos are attached as Supplementary Material. It can be seen that the tracker performs better in the generated videos as these are less challenging than the original ones which contain large changes in illumination and occlusion.  
\subsection{Robustness against landmark errors}
\noindent Finally, we evaluated the robustness of our model against errors in the given landmarks. In particular, for a set of $4000$ test images, we first slightly perturbed the ground-truth points using a small rotation angle, and then perturbed the points further according to a random noise, with standard deviation varying according to different values of $\sigma$. We computed the FID scores w.r.t. the original images. The results and a visual example, are shown in Table~\ref{tableii}. We can see that the method tries to place the landmarks where targeted, while maintaining the structure of a face. 
\begin{table}[]
\centering
\caption{Top: FID measured on $4000$ generated images. Bottom: Example results for each $\sigma$-controlled landmark noise.}
\label{tableii}
\begin{tabular}{|c|c|c|c|c|c|}
\hline
$\sigma$ & 1   & 2   & 3   & 4   & 5            \\ \hline
FID      & 0.9 & 2.4 & 4.5 & 7.7 & 13.8  \\ \hline
 {\includegraphics[width=0.12\linewidth]{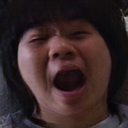}} & {\includegraphics[width=0.12\linewidth]{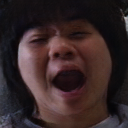}} &
 {\includegraphics[width=0.12\linewidth]{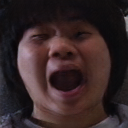}} &
 {\includegraphics[width=0.12\linewidth]{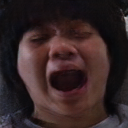}} &
 {\includegraphics[width=0.12\linewidth]{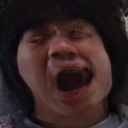}} &
 {\includegraphics[width=0.12\linewidth]{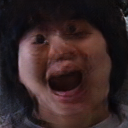}} \\ \hline
  {\includegraphics[width=0.12\linewidth]{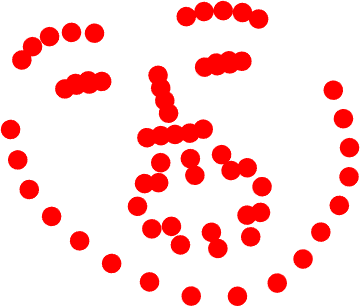}} & {\includegraphics[width=0.12\linewidth]{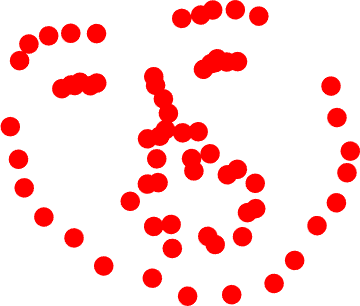}} &
 {\includegraphics[width=0.12\linewidth]{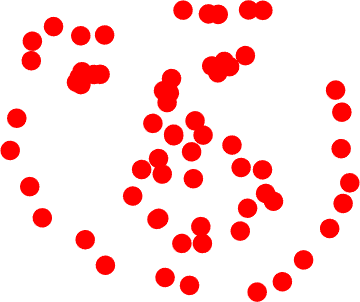}} &
 {\includegraphics[width=0.12\linewidth]{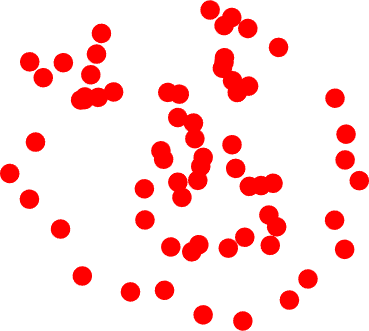}} &
 {\includegraphics[width=0.12\linewidth]{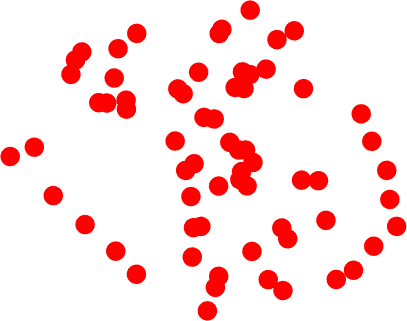}} &
 {\includegraphics[width=0.12\linewidth]{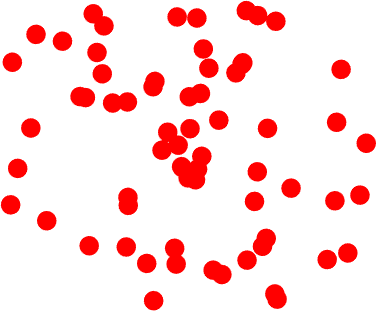}} \\ \hline
\end{tabular}
\end{table}
\section{Conclusion}
\noindent In this paper, we have illustrated a drawback of face-to-face synthesis methods that aim to preserve the input appearance by using a cycle consistency loss. We have shown that despite images being realistic, they cannot be reused by the network for further tasks. Based on this evidence, we have introduced a recurrent cycle consistency loss, which attempts to make the network reproduce similar results independently of the number of steps used to reach the target. We have incorporated this loss into a new landmark-guided face synthesis, coined GANnotation. We showed how the target landmarks become the ground-truth points, thus making GANnotation a powerful tool. We believe this paper opens the research question of making images reusable even when the results support plausible images. 
\section*{Acknowledgment}
\noindent This research was co-funded by the NIHR Nottingham Biomedical Research Centre.

{\small
\bibliographystyle{ieee}
\bibliography{egbib}
}

\end{document}